\pdfoutput=1
\documentclass[sigconf,nonacm]{acmart}

\usepackage{multirow}

\begin{document}

\title{Activity-Dependent Plasticity in Morphogenetically-Grown Recurrent Networks}

\author{Sergii Medvid}
\orcid{0009-0002-5601-4471}
\affiliation{%
  \institution{National University of Kyiv-Mohyla Academy}
  \city{Kyiv}
  \country{Ukraine}
}
\email{s.medvid@ukma.edu.ua}

\author{Andrii Valenia}
\orcid{0009-0004-4559-5500}
\affiliation{%
  \institution{National University of Kyiv-Mohyla Academy}
  \city{Kyiv}
  \country{Ukraine}
}
\email{a.valenia@ukma.edu.ua}

\author{Mykola Glybovets}
\orcid{0009-0005-6942-8026}
\affiliation{%
  \institution{National University of Kyiv-Mohyla Academy}
  \city{Kyiv}
  \country{Ukraine}
}
\email{glib@ukma.edu.ua}

\ccsdesc[500]{Computing methodologies~Neural networks}
\ccsdesc[300]{Computing methodologies~Bio-inspired approaches}

\begin{abstract}
Developmental neural architecture search grows functional networks from compact genomes via self-organisation, but the resulting weights are frozen post-growth---unlike biological circuits, which combine developmental wiring with activity-dependent plasticity.
Can post-developmental plasticity meaningfully improve such networks, and can evolution discover the right parameters per architecture?
We address these through population-scale characterisation on two control benchmarks (50{,}000 networks, 5M+ configurations), co-evolution of plasticity with the developmental genome, and a random-RNN control disentangling developmental from generic effects.
First, anti-Hebbian plasticity reliably outperforms Hebbian for competent networks (Cohen's $d = 0.53$--$0.64$).
Second, \emph{regret}---the fraction of oracle improvement lost under the best fixed setting---reaches 52--100\%, indicating per-network heterogeneity.
Third, under abrupt physics changes, plasticity shifts from fine-tuning to genuine within-lifetime adaptation, with an OFF$\to$ON protocol retaining 66--69\% of the always-on benefit.
Co-evolution rediscovers these patterns: 70\% of CartPole runs evolve anti-Hebbian ($p = 0.043$); Acrobot evolves near-zero $\eta$ with mixed signs.
A random-RNN control reveals that anti-Hebbian dominance is generic to small recurrent networks, but the \emph{degree} of topology-dependence is developmental-specific: regret is 2--6$\times$ higher in morphogenetic than topology-matched random architectures.
\end{abstract}

\keywords{developmental neural networks, synaptic plasticity, morphogenetic self-organisation, neuroevolution, indirect encoding}

\maketitle

\section{Introduction}
\label{sec:intro}

Developmental approaches to neural architecture search---including Neural Developmental Programs (NDP)~\cite{najarro2023ndp}, HyperNCA~\cite{najarro2022hypernca}, and MorphoNAS~\cite{glybovets2026morphonas}---encode compact genomes that unfold into functional neural networks through developmental processes. These systems demonstrate that self-organising developmental rules can produce effective controllers from far fewer parameters than direct weight specification, echoing the genomic bottleneck observed in biological brains~\cite{zador2019critique}.

In MorphoNAS, as in NDP and HyperNCA, the network topology and weights are fixed once development completes. This contrasts with biological neural circuits, where activity-dependent synaptic plasticity enables within-lifetime adaptation~\cite{hebb1949organization}. Recent work on Hebbian neuroevolution~\cite{najarro2020meta} has shown that evolving plasticity rules can produce adaptive agents, but this work operates on fixed-topology networks rather than developmentally grown architectures. Palm et~al.~\cite{palm2021testing} tested the genomic bottleneck hypothesis with Hebbian meta-learning but found that naive bottleneck application to plasticity rules leads to poor performance, suggesting a difficult joint optimisation problem.

Two fundamental questions arise. First, \textbf{can post-development\-al plasticity meaningfully improve morphogenetically grown networks, and if so, under what conditions?} Second, \textbf{can evolutionary optimisation discover the right plasticity parameters for each architecture, and does this interaction differ between developmental and random networks?} These questions are central to evolving self-organising systems: structural self-organisation (development) and functional self-organisation (plasticity) may interact in ways that neither uniform parameter tuning nor isolated analysis can reveal.

We present a systematic characterisation of synaptic plasticity in developmental networks at population scale, followed by co-evolutionary experiments that test whether evolutionary optimisation recovers the characterisation patterns. We use MorphoNAS~\cite{glybovets2026morphonas} as our experimental platform. Its reaction-diffusion dynamics~\cite{turing1952chemical} produce structurally diverse topologies from compact genomes, making it a suitable testbed for the broader class of developmental encodings. We evaluate Hebbian and anti-Hebbian plasticity (positive vs.\ negative learning rate $\eta$; anti-Hebbian weakens connections between co-active neurons rather than strengthening them) across 50{,}000 morphogenetically grown CartPole controllers under three task conditions: static, plus two non-stationary variants---10$\times$ pole mass and 2$\times$ gravity---where physics parameters change abruptly mid-episode without notification to the agent. We replicate key findings on 5{,}000 Acrobot controllers, then test whether co-evolving plasticity parameters alongside the developmental genome recovers the characterisation patterns. A random-RNN control disentangles developmental from generic network effects. Our contributions are:

\begin{itemize}
  \item A stratified characterisation across 3.75M evaluations revealing anti-Hebbian dominance (Cohen's $d = 0.53$--$0.64$), topology-dependent regret (52--100\%), and a functional role-shift from fine-tuning to genuine within-lifetime adaptation under non-stationarity.
  \item Co-evolutionary validation: when plasticity parameters are encoded in the genome and evolved, evolution independently discovers anti-Hebbian dominance on CartPole ($p = 0.043$) and the correct $\eta$ scale on both benchmarks (CartPole favours $|\eta| \sim 10^{-2}$; Acrobot needs $|\eta|$ roughly $10$--$100\times$ smaller).
  \item A random-RNN control showing that anti-Hebbian dominance is generic to small RNNs, but the \emph{degree} of topology-dependence (regret) is 2--6$\times$ higher for developmental networks: morphogenetic structure creates uniquely diverse plasticity interactions.
\end{itemize}

\section{Related Work}
\label{sec:related}

\textbf{Developmental neural networks.}
Indirect encodings that grow neural networks through developmental processes have a rich history in evolutionary computation. Stanley's NEAT~\cite{stanley2002evolving} complexifies topology through mutation, while HyperNEAT~\cite{stanley2009hypercube} uses compositional pattern-producing networks to exploit geometric regularities. Neural Cellular Automata (NCA)~\cite{mordvintsev2020growing} demonstrated that self-organising update rules can grow complex patterns from local interactions; Neural Developmental Programs (NDP)~\cite{najarro2023ndp} and HyperNCA~\cite{najarro2022hypernca} extend this to grow functional neural networks through self-assembling developmental processes. MorphoNAS~\cite{glybovets2026morphonas} takes a complementary developmental path, using Turing-type reaction-diffusion morphogen dynamics~\cite{turing1952chemical} rather than learned update rules. Nisioti et~al.~\cite{nisioti2024growing} showed that maintaining neuronal diversity during NDP growth---via intrinsic hidden states and lateral inhibition---is essential for producing functional controllers, providing independent evidence that the developmental process shapes the structural heterogeneity of the resulting networks. In all these systems, weights are typically fixed during task execution.

\textbf{Plasticity in neuroevolution.}
Soltoggio et~al.~\cite{soltoggio2018born} survey the field of evolved plastic networks, identifying Hebbian and neuromodulated plasticity as the dominant paradigms. Najarro and Risi~\cite{najarro2020meta} demonstrated that Hebbian learning rules evolved via meta-learning can produce agents that adapt within their lifetime---but on fixed-topology random networks, not developmentally grown ones. Plantec et~al.~\cite{plantec2024evolving} introduced the Lifelong Neural Developmental Program (LNDP), combining graph-transformer-based development with experience-dependent structural plasticity, enabling networks to add and prune connections during task execution. Garc{\'\i}a~N{\'u}{\~n}ez et~al.~\cite{garcianunez2025time} proposed time-modulated Hebbian learning (tm-HL) for classic control tasks, introducing a decay term $\mu_t = e^{-\lambda t}$ that transitions networks from high plasticity to fixed weights over the agent's lifetime; they used $\lambda = 0.01$---the same value we find effective for weight decay. However, their work uses fixed feedforward architectures and does not test anti-Hebbian rules. Our work complements theirs by characterising plasticity across 50{,}000 \emph{developmentally grown} topologies, revealing that the interaction between plasticity and network structure---rather than plasticity parameters alone---is the dominant factor.

\textbf{The genomic bottleneck.}
Zador~\cite{zador2019critique} argued that biological brains leverage compact genomic encodings that specify wiring rules rather than individual weights, with experience fine-tuning pre-wired circuits. Palm et~al.~\cite{palm2021testing} tested whether a genomic bottleneck improves Hebbian meta-learning but found the joint optimisation problem difficult under their approach. Our work takes a different approach: we first characterise how simple plasticity interacts with the topology already compressed through morphogenetic development, then test whether encoding plasticity parameters in the genome and co-evolving them alongside the developmental architecture can exploit this interaction.

\section{Experimental Design}
\label{sec:design}

\subsection{MorphoNAS Developmental System}

\begin{figure}[t]
  \centering
  \includegraphics[width=\columnwidth]{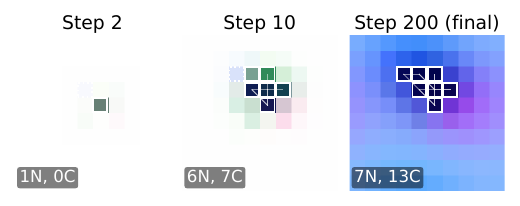}
  \Description{Three grid panels labelled Step 2, Step 10, and Step 200 (final), showing a developing MorphoNAS network. Each grid uses a coloured background to encode three diffusing morphogen concentrations as RGB channels: pale at Step 2 (one neuron, no connections), mixed greens and pinks at Step 10 (6 neurons, 7 connections), saturated purple-blue gradient at Step 200 (7 neurons, 13 connections). White squares mark neurons; white lines mark connections.}
  \caption{Developmental dynamics of a MorphoNAS network on a $10 \times 10$ grid. Background colour encodes morphogen concentrations (RGB = 3 morphogens); white squares = neurons; white lines = connections. From a single progenitor cell (Step~2), reaction-diffusion dynamics drive cell division, differentiation, and chemotactic axon growth, producing a 7-neuron, 13-connection recurrent controller by Step~200.}
  \label{fig:morphonas}
\end{figure}

We use MorphoNAS~\cite{glybovets2026morphonas}, a morphogenetic neural architecture search system where a compact genome (54 parameters for 3 morphogens) encodes morphogen secretion rates, diffusion tensors, cross-inhibi\-tion coefficients, cell-fate thresholds, and axon growth rules. Starting from a single progenitor cell on a 2D toroidal grid, each developmental iteration involves morphogen secretion, spatial diffusion, competitive cross-inhibition (producing reaction-diffusion dynamics and symmetry breaking), local cell-fate decisions (division, differentiation, or quiescence), and chemotactic axon growth with distance-dependent weight initialisation ($w = \max(0.01,\; c/(1+d))$). After a genome-specified number of iterations, the process yields a recurrent neural network (RNN) whose topology and weights emerge entirely from local chemical interactions (Figure~\ref{fig:morphonas}). The process is fully deterministic: the same genome always produces the same network. Crucially, the 54 parameters do not specify individual neurons or connections---they encode morphogen dynamics whose self-organising interactions constrain the phenotype space to morphogenetically plausible topologies. Different genomes produce structurally distinct networks, and it is these structural differences that make the interaction with plasticity topology-dependent (Section~\ref{sec:results}). See~\cite{glybovets2026morphonas} for full details.

\subsection{Task and Network Pool}

The resulting networks are evaluated as controllers on CartPole-v1 (OpenAI Gymnasium~\cite{towers2024gymnasium}), where the objective is to balance a pole on a cart for up to 500~timesteps. CartPole is deliberately chosen as a minimal substrate: its simplicity isolates the development-plasticity interaction from task complexity, while mid-episode physics perturbations (Section~\ref{sec:adaptation}) reveal how plasticity's role changes with environmental demands.

We generated 50{,}000 networks from \emph{randomly sampled} genomes (no evolutionary selection; each genome drawn independently from the MorphoNAS parameter space). We stratified networks by baseline CartPole performance (mean reward over 20~episodes without plasticity):

\begin{itemize}
  \item \textbf{Weak} ($r < 200$): $N = 47{,}638$ (95.3\%)
  \item \textbf{Low-mid} ($200 \le r < 350$): $N = 1{,}235$
  \item \textbf{High-mid} ($350 \le r < 450$): $N = 256$
  \item \textbf{Near-perfect} ($450 \le r < 475$): $N = 102$
  \item \textbf{Perfect} ($r \ge 475$): $N = 769$
\end{itemize}

That 95.3\% of random genomes produce Weak networks indicates that, in our setting, \emph{viable topology is the primary determinant of performance}: while 15\% of Weak networks show marginal improvement, the developmental process must produce a functional architecture before plasticity can contribute substantially.

\subsection{Plasticity Protocol}

We apply a Hebbian plasticity rule after development completes. At each propagation timestep during task execution, weights are updated according to:
\begin{equation}
  \Delta w_{ij} = \eta \cdot x_i \cdot x_j - \lambda \cdot w_{ij}
  \label{eq:hebb}
\end{equation}
where $\eta$ is the learning rate (positive for Hebbian, negative for anti-Hebbian), $x_i, x_j$ are pre- and post-synaptic activations, and $\lambda$ is a weight decay coefficient that acts as a stability mechanism.

Each network is evaluated under a rigorous protocol: (1)~baseline measurement with plasticity disabled, (2)~plasticity-enabled evaluation over 20~episodes with fixed seeds (42--61), (3)~fitness comparison. The improvement $\Delta r$ is the difference in mean reward between the plasticity-enabled and baseline evaluations. All experiments ran on AWS c6i.4xlarge instances (16~vCPU, Intel Xeon 8375C). The full experimental programme totals over 5~million plasticity configurations across CartPole and Acrobot, plus 180 evolutionary runs and 11{,}810 random-graph controls, requiring approximately 7{,}700 core-hours. Per-experiment grids and sample sizes are presented alongside their results in Sections~\ref{sec:results} and~\ref{sec:evolution}.

\section{Characterisation Results}
\label{sec:results}

\subsection{Stratified Plasticity Impact}
\label{sec:stratified}

We characterised the plasticity landscape across the full pool of 50{,}000 networks on a 75-point grid ($\eta \in [-0.05, +0.05]$, 15 non-uniform levels; $\lambda \in \{0, 10^{-5}, 10^{-4}, 10^{-3}, 10^{-2}\}$), totalling 3{,}750{,}000 evaluations. Table~\ref{tab:stratified} presents the per-stratum impact of plasticity under \emph{oracle} tuning---exhaustive per-network grid search for the best $(\eta, \lambda)$ pair, representing an upper bound on achievable benefit. We define \emph{regret} as the fraction of per-network oracle improvement lost when committing to a single best fixed $(\eta, \lambda)$ for the entire population; high regret means parameter tuning must be per-network rather than universal.

\begin{table}[t]
\caption{CartPole plasticity impact by performance stratum under per-network oracle parameters; regret as defined in the text.}
\label{tab:stratified}
\small
\begin{tabular}{@{}lrrrrr@{}}
\toprule
Stratum & $N$ & Oracle $\Delta r$ & \% Impr. & Best $\eta$ & Regret \\
\midrule
Weak       & 47{,}638 & $+2.2$   & 15\% & $+0.05$ & 59\%  \\
Low-mid    & 1{,}235  & $+35.7$  & 50\% & $-0.05$ & 52\%  \\
\textbf{High-mid} & \textbf{256} & $\mathbf{+60.6}$ & \textbf{93\%} & $\mathbf{-0.05}$ & \textbf{86\%} \\
Near-perf. & 102      & $+22.5$  & 84\% & $-0.005$ & 89\% \\
Perfect    & 769      & $+7.7$   & 20\% & ---     & 100\% \\
\bottomrule
\end{tabular}
\end{table}

Three key patterns emerge. First, plasticity \emph{cannot rescue weak topologies}: networks with non-functional architectures show negligible improvement regardless of plasticity parameters. In our experiments, topology produced by morphogenetic development is the primary determinant of viability. Second, \emph{competent networks benefit substantially}: up to 93\% of High-mid networks improve under oracle tuning, with mean gains of +60.6 reward points. Third, \emph{per-network parameter heterogeneity is large}: regret under fixed parameters ranges from 52\% (Low-mid) to 100\% (Perfect), indicating that no single parameter setting captures the oracle potential.

Table~\ref{tab:stratified} reports conservative estimates from the 75-point primary sweep; the best $\eta$ for Low-mid and High-mid falls at the grid boundary ($|\eta| = 0.05$). The extended 248-point validation grid, which widens the range to $|\eta| \le 0.5$, yields higher oracle improvements: $+73.4$ for Low-mid and $+86.3$ for High-mid. Split-half cross-validation (oracle selected on odd-numbered episodes, evaluated on even) confirmed that 74--92\% of the oracle advantage reflects genuine per-network heterogeneity, not overfitting to noise. Under ceiling-corrected normalisation (gain as fraction of available headroom), High-mid networks capture 87\% of their improvement room and Near-perfect 78\%, indicating that oracle plasticity nearly saturates the performance gap.

The magnitude of oracle gains enables \emph{stratum transitions}: under per-network tuning, a majority of High-mid and Near-perfect networks reach the Perfect stratum ($r \ge 475$), suggesting that plasticity can close the gap between ``competent but imperfect'' and ``fully functional'' controllers---a qualitative shift in behaviour, not merely a marginal improvement.

However, plasticity carries asymmetric risk. Harm rates increase with baseline competence: 1.3\% (Weak), 18.5\% (Low-mid), 42.0\% (High-mid), 48.5\% (Near-perfect). The worst-case configurations ($\eta > 0$, $\lambda = 0$) produce severe degradation, with mean normalised harm of $-0.17$ to $-0.24$ ($-85$ to $-120$ raw reward points) for competent strata. This asymmetry---where the networks most likely to benefit are also most vulnerable to harm---further motivates per-network parameter selection over fixed configurations.

\subsection{Anti-Hebbian Dominance}

To resolve the boundary effect at $|\eta| = 0.05$ noted above, we ran an extended grid on 2{,}862 non-Weak networks (including borderline cases near stratum boundaries) with $\eta \in [-0.5, +0.5]$ across 31 levels and $\lambda \in [0, 0.1]$ across 8 levels (248 grid points, 709{,}776 evaluations). This grid underpins the cross-validated regret estimates above and the anti-Hebbian effect sizes below.

\begin{figure}[t]
  \centering
  \includegraphics[width=\columnwidth]{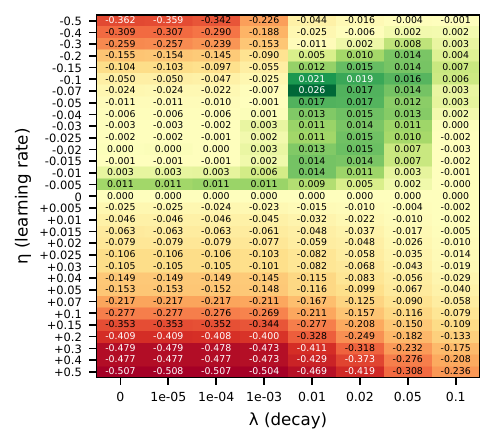}
  \Description{Heatmap of mean reward improvement on a 248-point grid: decay lambda on the x-axis (8 values from 0 to 0.1) and learning rate eta on the y-axis (31 values from -0.5 at the top to +0.5 at the bottom). Each cell shows a numerical delta-r value with a diverging colour scale: warm tones (red) for negative values where plasticity hurts, cool tones for positive values where plasticity helps. Highest positive values cluster around eta from -0.1 to -0.005 with decay 0.01-0.02; the most negative values appear at the |eta| > 0.2 extremes in either direction.}
  \caption{CartPole High-mid stratum: mean $\Delta r$ across the $\eta \times \lambda$ grid (248-point extended grid). Anti-Hebbian ($\eta < 0$) with moderate decay ($\lambda = 0.01$) yields the best results. Hebbian ($\eta > 0$) is consistently harmful.}
  \label{fig:heatmap}
\end{figure}

\begin{figure}[t]
  \centering
  \includegraphics[width=\columnwidth]{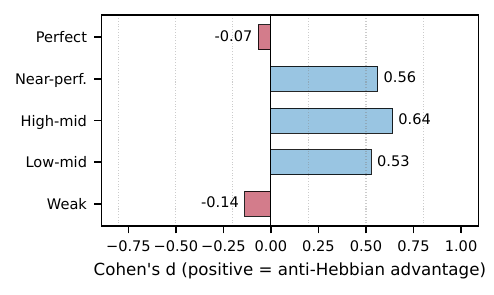}
  \Description{Horizontal bar chart of Cohen's d (positive = anti-Hebbian advantage) across five performance strata listed top to bottom as Perfect, Near-perf., High-mid, Low-mid, Weak. Three blue bars extend rightward: Near-perf. 0.56, High-mid 0.64, Low-mid 0.53. Two short pink bars extend leftward: Perfect -0.07, Weak -0.14. The horizontal axis spans from -0.75 to +1.00.}
  \caption{CartPole: Cohen's $d$ (anti-Hebbian $-$ Hebbian) by stratum. Anti-Hebbian significantly outperforms Hebbian for competent strata ($d = 0.53$--$0.64$, all $p < 0.001$). The effect reverses for Weak and Perfect networks with small magnitudes.}
  \label{fig:cohens_d}
\end{figure}

Anti-Hebbian plasticity ($\eta < 0$) significantly outperforms Hebbian ($\eta > 0$) for all competent strata (Figure~\ref{fig:cohens_d}). On the extended 248-point grid, the effect is substantial: Cohen's $d = 0.53$ for Low-mid, $d = 0.64$ for High-mid, and $d = 0.56$ for Near-perfect (all $p < 0.001$). The effect reverses for Weak ($d = -0.14$) and Perfect ($d = -0.07$) networks, though with small absolute magnitudes.

Kruskal-Wallis variance decomposition on the extended parameter grid~\cite{derrac2011practical} reveals that learning rate ($\eta$) explains 22--35\% of variance in competent strata ($\varepsilon^2 = H/(N{-}1)$), while decay ($\lambda$) explains less than 5\%. This indicates that \emph{the sign and magnitude of $\eta$ is the critical design choice}, while decay acts primarily as a safety mechanism. The sweet spot of $\lambda = 0.01$ reduces harm rates by 3--8$\times$ compared to $\lambda = 0$ with minimal reduction in peak benefit.

What drives the anti-Hebbian advantage? We tested two mechanistic hypotheses. First, that anti-Hebbian plasticity acts as a decorrelation mechanism; we found no relationship between initial weight magnitude and anti-Hebbian preference ($p = 0.56$, Mann-Whitney). Second, that Hebbian updates overshoot by reinforcing existing correlations too aggressively; correlational evidence is consistent: networks whose optimal $\eta$ is positive (Hebbian) exhibit 2.6$\times$ larger per-step weight changes than anti-Hebbian-preferring networks ($p < 10^{-25}$; per-stratum $d = 0.34$--$0.87$). The lowest weight-change quintile shows 93\% anti-Hebbian preference versus 64\% in the highest quintile. While not dispositive, this pattern is consistent with \emph{magnitude moderation}---anti-Hebbian updates counteract co-activation rather than amplifying it, preventing the runaway weight growth that makes Hebbian updates harmful for competent networks.

\subsection{Within-Lifetime Adaptation}
\label{sec:adaptation}

To test whether plasticity can serve a qualitatively different role than static fine-tuning---genuine adaptation to changed dynamics---we evaluated 2{,}362 non-Weak networks on two non-stationary CartPole variants where physics change abruptly at timestep~200 with no signal to the agent: pole mass $\times$10 (0.1$\to$1.0\,kg, affecting inertial dynamics) and gravity $\times$2 (9.8$\to$20.0\,m/s$^2$, uniformly scaling forces). Each variant used a 22-point anti-Hebbian-range grid (103{,}928 total evaluations).

\begin{figure}[t]
  \centering
  \includegraphics[width=\columnwidth]{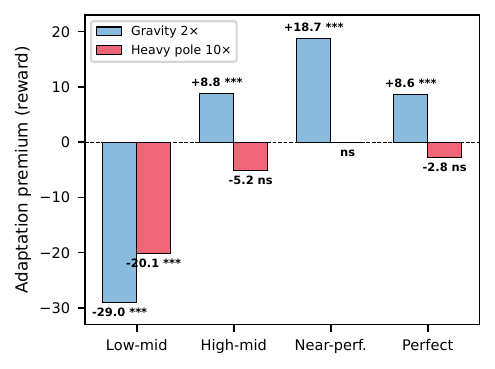}
  \Description{Vertical grouped bar chart of CartPole adaptation premium (reward, y-axis from -30 to +20) across four performance strata on the x-axis: Low-mid, High-mid, Near-perf., Perfect. Each stratum has two bars: blue for Gravity 2x, pink for Heavy pole 10x. Low-mid is strongly negative for both variants (-29.0 *** blue and -20.1 *** pink). High-mid through Perfect show positive significant Gravity 2x premiums (+8.8, +18.7, +8.6, all ***) and small or non-significant Heavy pole 10x values (-5.2 ns, ns, -2.8 ns).}
  \caption{CartPole adaptation premium (non-stationary plasticity benefit minus static plasticity benefit) for two perturbation types. Gravity-2$\times$ yields significant positive premiums for competent strata (***$p<0.001$), indicating genuine adaptation beyond static fine-tuning. Heavy-pole shows near-zero premiums (n.s.).}
  \label{fig:adaptation}
\end{figure}

Under both variants, 88--90\% of High-mid networks benefit from plasticity in Phase~2 (post-perturbation). The gravity-2$\times$ variant yields a significant \emph{adaptation premium} (non-stationary benefit minus static benefit): $+8.8$ to $+18.7$ mean reward for competent strata ($p < 0.001$, Wilcoxon; Figure~\ref{fig:adaptation}), while heavy-pole shows near-zero premiums. Weight-change dynamics differ accordingly: under gravity-2$\times$, mean $|\Delta w|$ increases post-switch (ratio 1.13), consistent with active adaptation; under heavy-pole, it decreases (ratio 0.82--0.93), suggesting robustness rather than new adaptation.

\textbf{Causal confirmation via OFF$\to$ON.} To isolate adaptation from robustness, we disabled plasticity during Phase~1 and enabled it only at the switch point (103{,}928 OFF$\to$ON evaluations on the same 2{,}362 networks). OFF$\to$ON retains 66--69\% of the always-on benefit: 84--86\% of High-mid networks still improve, with oracle gains of $+53.9$ (heavy-pole) and $+59.9$ (gravity-2$\times$), representing \emph{pure} within-lifetime adaptation. A dose-response experiment varying switch time (100--400 steps under heavy-pole; 51{,}964 evaluations) confirms: benefit scales monotonically with post-switch duration (Spearman $\rho = 1.0$, all strata), from $+6.1$ at 100~steps to $+74.2$ at 400.

\subsection{Replication on Acrobot}
\label{sec:acrobot}

To test whether these findings generalise beyond CartPole, we repeated the analysis on Acrobot-v1, a harder control task with 6-dimensional observations and 3 discrete actions. We sampled 5{,}000 random genomes on a $20 \times 20$ developmental grid; 89.7\% were non-functional, leaving 513 networks across four performance strata, of which 87 (1.7\%) meet Gymnasium's solved threshold. We evaluated these networks on a combined parameter grid---coarse 22-point sweep plus an extended 248-point micro-$\eta$ grid ($|\eta| \in [5 \times 10^{-5}, 0.1]$, $\lambda \in [0, 0.1]$)---under both static and non-stationary conditions (2$\times$ lower-link mass at step~50). Under per-network oracle tuning, 93--97\% of networks per stratum improve with plasticity (Table~\ref{tab:acrobot}). Acrobot requires substantially smaller learning rates than CartPole: the optimal $|\eta|$ is $10^{-4}$--$10^{-3}$ with high decay ($\lambda = 0.05$--$0.1$), roughly 10--100$\times$ smaller than CartPole's sweet spot. Yet regret under any fixed parameter is 100\%: no single $(\eta, \lambda)$ improves the average network, even on the extended grid. Plasticity is beneficial but entirely topology-dependent.

\begin{figure}[t]
  \centering
  \includegraphics[width=\columnwidth]{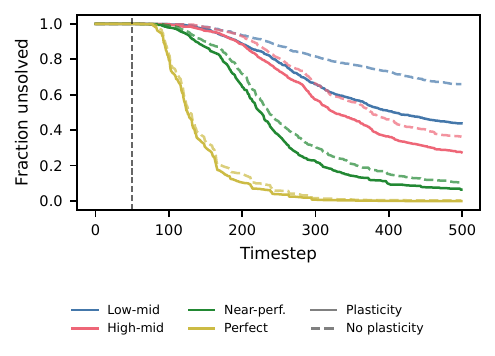}
  \Description{Single-panel plot of fraction-of-episodes-unsolved (y-axis 0 to 1) over timestep (x-axis 0 to 500) for Acrobot under non-stationarity. A vertical dashed black line at timestep 50 marks the perturbation switch (2x lower-link mass). Four colour-coded stratum curves: Low-mid (blue), High-mid (pink), Near-perf. (green), Perfect (yellow). Solid lines (Plasticity) drop faster than dashed lines (No plasticity) across all strata, with the largest separation in the High-mid and Near-perfect groups.}
  \caption{Acrobot fraction of episodes unsolved over time under non-stationarity (2$\times$ link mass at step~50). Solid: per-network oracle plasticity; dashed: no plasticity. Lower is better. Oracle plasticity accelerates solving across all strata---most visibly for High-mid and Near-perfect, where the unsolved fraction at step~300 drops from 66\%/31\% (no plasticity) to 40\%/18\% (oracle plasticity).}
  \label{fig:acrobot_survival}
\end{figure}

The anti-Hebbian picture is more nuanced on Acrobot (Table~\ref{tab:acrobot}). Cohen's $d$ favours anti-Hebbian across all strata ($d = 0.09$--$0.48$), but the per-network optimal $\eta$ splits roughly evenly between signs (151 anti-Hebbian vs 185 Hebbian). Under non-stationarity (2$\times$ lower-link mass at step~50) with the combined grid, 87--96\% of networks per stratum benefit from per-network oracle plasticity---again with 100\% regret under any fixed setting. Plasticity accelerates solving across all strata (Figure~\ref{fig:acrobot_survival}): the fraction of unsolved episodes at step~300 drops from 66\% to 40\% (High-mid) and from 31\% to 18\% (Near-perfect) compared to the no-plasticity baseline. These results show that the development-plasticity interaction is not benchmark-specific: topology-dependent plasticity holds on a task with richer dynamics, though the non-stationary benefit requires per-network tuning on the current grid.

\begin{table}[t]
\caption{Acrobot-v1 plasticity characterisation. \% Impr.: per-network oracle improvement rate on the combined grid (coarse + micro-$\eta$). Cohen's $d$: anti-Hebbian vs Hebbian (coarse grid). NS: non-stationary oracle rate (combined grid). Regret is 100\% under any fixed $(\eta, \lambda)$.}
\label{tab:acrobot}
\small
\begin{tabular}{@{}lrrrr@{}}
\toprule
Stratum & $N$ & \% Impr. & Cohen's $d$ & NS \% Helped \\
\midrule
Low-mid       & 71  & 95.8\% & $+0.48$*** & 88.7\% \\
High-mid      & 74  & 93.2\% & $+0.31$*** & 95.9\% \\
Near-perf.    & 281 & 93.6\% & $+0.15$*** & 86.8\% \\
Perfect       & 87  & 96.6\% & $+0.09$*   & 95.4\% \\
\bottomrule
\multicolumn{5}{@{}l@{}}{\footnotesize ***$p<0.001$, *$p<0.05$}
\end{tabular}
\end{table}

\section{Evolutionary Validation and Controls}
\label{sec:evolution}

\subsection{Co-Evolution Recovers Characterisation Findings}
\label{sec:coevol_results}

The characterisation established three patterns testable by independent search: anti-Hebbian dominance on CartPole, near-zero $\eta$ with mixed signs on Acrobot, and topology-dependent regret (52--100\%). We now ask whether evolutionary optimisation recovers these without prior knowledge. We ran a genetic algorithm evolving MorphoNAS genomes (CartPole on $10 \times 10$, Acrobot on $20 \times 20$) under three conditions, with 30 independent runs each (population~50, 200~generations, top-20\% selection, 2-elitism, mutation rate~0.3): \textbf{A}~(no plasticity, frozen weights), \textbf{B}~(fixed plasticity at the characterisation defaults: $\eta = -0.01, \lambda = 0.01$ on CartPole; $\eta = -0.001, \lambda = 0.05$ on Acrobot), and \textbf{C}~(co-evolved plasticity, with two extra genome parameters mutating alongside the developmental genome; CartPole: $\eta \in [-0.5, +0.5]$, $\lambda \in [0, 0.1]$; Acrobot: $\eta \in [-0.1, +0.1]$, $\lambda \in [0, 0.1]$). Total: 180 GA runs of 10{,}050 fitness evaluations each.

\begin{figure}[t]
  \centering
  \includegraphics[width=\columnwidth]{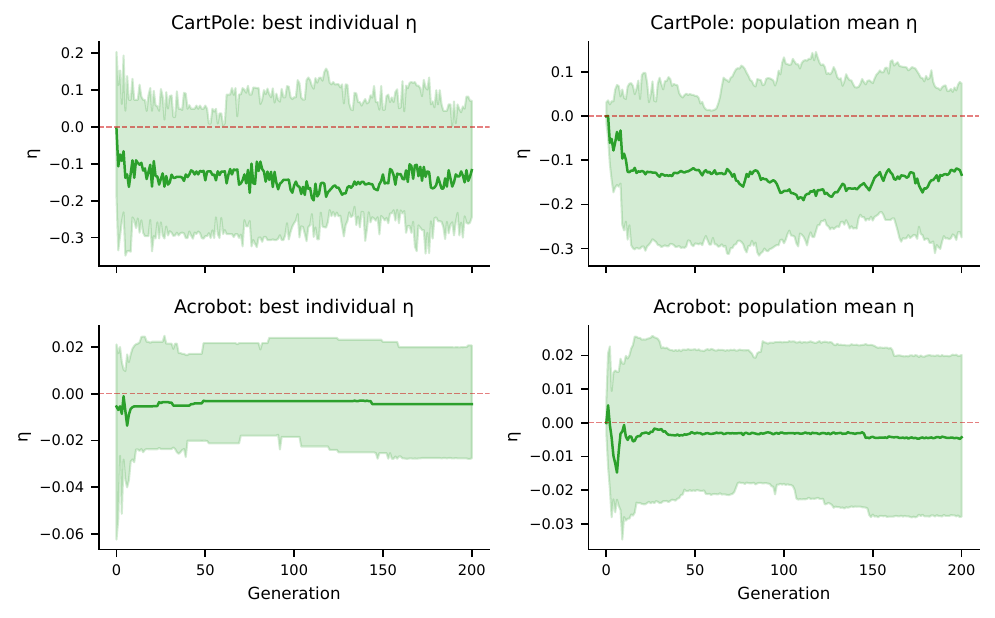}
  \Description{Four-panel plot (2 rows by 2 columns) of evolved learning rate eta over 200 GA generations. Top row: CartPole; bottom row: Acrobot. Left column: best individual eta per generation; right column: population mean eta with median (solid green line) and interquartile range (shaded band) across 30 runs. A red dashed reference line at eta = 0 separates Hebbian (positive) from anti-Hebbian (negative). CartPole panels show clear drift to eta around -0.1 to -0.2 (anti-Hebbian); Acrobot panels show eta fluctuating very close to zero with small mixed-sign deviations.}
  \caption{Co-evolution (Condition~C) evolved $\eta$ over 200 generations on CartPole (top) and Acrobot (bottom). Left: best individual's $\eta$; right: population mean $\eta$ (median $\pm$ IQR across 30 runs). CartPole: strong anti-Hebbian drift ($\eta < 0$). Acrobot: $\eta$ hovers near zero with mixed signs, matching the characterisation finding.}
  \label{fig:eta_evolution}
\end{figure}

On CartPole, all three conditions converge to perfect fitness (reward~500) within 1--2 generations---the benchmark is too easy to differentiate conditions on final fitness ($p = 1.0$, Mann-Whitney). However, the evolved plasticity parameters reveal a clear signal. Starting from a uniform prior over $[-0.5, +0.5]$, \textbf{21/30 (70\%) of Condition~C runs evolve negative $\eta$} (binomial sign test $p = 0.043$), with median evolved $\eta = -0.117$ and population-mean $\eta$ drifting below zero within $\sim$10 generations (Figure~\ref{fig:eta_evolution}). Evolution independently discovers anti-Hebbian dominance. Co-evolution also produces sparser networks than no-plasticity evolution (median 277 vs 328 connections, $p = 0.10$, Mann-Whitney; neuron count identical at 100). The trend is stronger against fixed-plasticity evolution (277 vs 357, $p = 0.004$), hinting at a Baldwin-like effect where lifetime plasticity relaxes structural requirements.

On static Acrobot, no condition breaks through the architectural ceiling: all plateau at fitness~$\approx 0.84$ (reward~$\approx -78$), with all 90 runs in the High-mid stratum ($p \ge 0.72$, all pairwise comparisons). The critical finding is in the evolved $\eta$ distribution, which is \textbf{qualitatively different from CartPole}: median $\eta = -0.0045$, mean $\approx 0$, with 18/30 (60\%) negative ($p = 0.36$, n.s.). Evolved $|\eta|$ clusters in $[10^{-3}, 5 \times 10^{-2}]$---the Acrobot sweet spot from the characterisation (Section~\ref{sec:acrobot}). The cross-task contrast is itself informative: evolution finds different answers for different tasks, exactly as the characterisation predicts.

We tested non-stationary co-evolution on Acrobot rather than CartPole: CartPole's fitness ceiling saturates within 1--2 generations under any condition (above), leaving no headroom to discriminate plasticity contributions, whereas Acrobot's lower fitness ceiling and tighter $\eta$-scale requirement (Section~\ref{sec:acrobot}) provide the comparison space in which architecture-vs-plasticity contributions can be separated. Under non-stationary Acrobot (2$\times$ lower-link mass at step~50), co-evolu\-tion is \textbf{counterproductive}: Condition~C (median fitness 0.775) is marginally worse than Condition~A (0.786; $d = 0.60$, $p = 0.033$ uncorrected, $p = 0.10$ after Bonferroni) and converges $3\times$ slower (median generation 10 vs 3 to reach fitness~0.75). The larger search space (56 vs 54 parameters) slows exploration without compensating fitness gains. The GA finds topologies that handle the perturbation through structure alone---Condition~A matches Condition~B---confirming that evolutionary architecture search absorbs non-stationarity without needing plasticity.

\subsection{Random-RNN Control}
\label{sec:random_results}

To disentangle developmental from generic network effects, we generated 11{,}810 random directed graphs matched to the topology statistics (neuron count, connection count) of 2{,}362 competent MorphoNAS CartPole networks (5~random RNNs per source network), with weights drawn from $\text{Uniform}[0.01, 1.0]$ to match MorphoNAS's range without distance-dependent structure. We applied the same 75-point plasticity grid as the primary characterisation sweep. The control yields three results.

\textbf{Developmental competence advantage.} Of 11{,}810 random RNNs matched to competent MorphoNAS topologies, only 65 (0.56\%) are themselves competent (Low-mid: 17, High-mid: 18, Near-perfect: 4, Perfect: 26), compared to MorphoNAS's 4.7\%---an $8.4\times$ difference. Morphogenetic development provides a genuine structural inductive bias beyond topology statistics alone.

\textbf{Anti-Hebbian dominance is generic.} Anti-Hebbian plasticity dominates for competent random RNNs \emph{even more strongly} than for MorphoNAS networks: Cohen's $d = 1.67$ [95\% bootstrap CI: 1.57, 1.78] for High-mid and $d = 1.79$ [1.69, 1.90] for Low-mid, vs $d = 0.64$ and $d = 0.53$ for MorphoNAS. The magnitude-moderation mechanism operates regardless of how the network was wired; anti-Hebbian dominance is a property of small recurrent networks on CartPole, not of developmental topology specifically.

\textbf{Topology-dependence \emph{is} developmental-specific.} Regret under fixed parameters is substantially higher for MorphoNAS networks: 86\% (High-mid) vs 40\% for random RNNs, and 89\% (Near-perfect) vs 14\%. Morphogenetic development creates more diverse topology-plasticity interactions, making per-network tuning---or co-evolution---more important for developmental systems than for random architectures.

\section{Discussion and Conclusion}
\label{sec:discussion}

The combination of characterisation, co-evolutionary validation, and random-RNN control allows us to separate general from deve\-lopmental-specific findings with unusual precision.

\textbf{Separating general from developmental-specific findings.} The random-RNN control (Section~\ref{sec:random_results}) resolves a question that the characterisation alone could not answer. Anti-Hebbian dominance turns out to be a generic property of small recurrent networks---the magnitude-moderation mechanism operates regardless of wiring origin. This recontextualises the characterisation finding: practitioners should default to anti-Hebbian plasticity not because of anything specific to developmental topology, but because it prevents runaway weight growth in any small RNN. By contrast, the \emph{degree} of topology-dependence is genuinely developmental: the $2$--$6\times$ higher regret for MorphoNAS networks means that morphogenetic development creates a uniquely diverse phenotype space where per-network tuning (or co-evolution) is far more important than for random architectures.

\textbf{Co-evolution recovers characterisation patterns but does not improve fitness.} On no benchmark or condition does co-evolution improve fitness over no-plasticity evolution (Section~\ref{sec:coevol_results}). Under non-stationary Acrobot, co-evolution is marginally \emph{worse} ($d = 0.60$): the two extra parameters slow convergence without payoff, because the GA finds topologies robust to the perturbation through structure alone. This reinforces a central finding: \emph{topology is the primary determinant of performance}---plasticity improves a given topology (characterisation) but does not help the evolutionary search for topology. The scientific contribution is methodological: evolution independently recovers the characterisation findings (anti-Hebbian on CartPole, mixed-sign near-zero on Acrobot, correct $\eta$ scale) without prior knowledge. The $8.4\times$ developmental competence advantage (Section~\ref{sec:random_results}) further underscores that reaction-diffusion development provides structural inductive bias beyond topology statistics alone.

\textbf{Architecture search vs.\ within-lifetime adaptation.} Our negative co-evolution result under non-stationarity is consistent with a \emph{learning-not-to-learn} hypothesis from meta-RL theory~\cite{lange2022learning}: when the cost of discovering an adaptive strategy within a fixed lifetime exceeds its benefit, the optimal solution is to hard-code a robust heuristic rather than learn one. Lange and Sprekeler show analytically that meta-learning systems exhibit two distinct regimes---learning and not-learning---separated by a sharp threshold determined by ecological uncertainty, task complexity, and expected lifetime; non-adaptive solutions can win even when adaptation would be highly beneficial, simply because it cannot be discovered and exploited fast enough. Our co-evolution result sits on the not-learning side of this threshold: within 200 generations on non-stationary Acrobot, the GA discovers structurally robust topologies more readily than it discovers \emph{and} exploits beneficial plasticity, even though the characterisation proves such plasticity exists for these topologies. This raises a methodological question for evolving self-organising systems: \emph{does evolutionary architecture search systematically bias toward robustness over adaptability?} Distinguishing ``the GA did not need plasticity'' from ``the GA could not exploit plasticity within 200 generations'' requires longer evolutionary horizons or harder benchmarks where structural robustness alone is insufficient---a direction we leave for future work.

\textbf{Implications for evolving self-organisation.} Our findings connect structural self-organisation (development) with functional self-organisation (plasticity) through evolution. In NDP~\cite{najarro2023ndp} and HyperNCA~\cite{najarro2022hypernca}, development is driven by learned update rules; Plantec et~al.~\cite{plantec2024evolving} showed that structural plasticity can extend development into the lifetime. Our results demonstrate that even simple synaptic plasticity provides substantial benefits on top of morphogenetic development, and that evolutionary optimisation can discover the appropriate plasticity parameters for each task. The sparser architectures evolved under co-evolution on CartPole (277 vs 357 connections compared to fixed-plasticity evolution, $p = 0.004$; Section~\ref{sec:coevol_results}) suggest a Baldwin-like effect where lifetime plasticity relaxes the structural requirements on the genome---a direction worth pursuing on harder benchmarks.

\textbf{Practical guidelines.} (1)~Default to anti-Hebbian plasticity ($\eta < 0$) with moderate decay ($\lambda \approx 0.01$)---this is both the characterisation recommendation and what evolution discovers independently. (2)~Invest in per-network parameter tuning: regret under fixed parameters is 52--100\% for MorphoNAS networks, far higher than for random architectures. (3)~When targeting non-stationary environments, the OFF$\to$ON protocol shows that adaptation alone accounts for 66--69\% of the total benefit.

\textbf{Benchmark limitations.} Both benchmarks are classic control tasks; generalisation to continuous-action or high-dimensional environments remains open. In particular, we did not test sparse-reward tasks such as MountainCar. Architecture search may be more susceptible to local optima under sparse rewards, where rare reward signals provide a weaker selection signal for discriminating between candidate topologies. This is an orthogonal challenge to the development-plasticity interaction we characterise---but it would be the natural next test of whether co-evolution can recover the per-network plasticity benefits we observe under dense rewards. We leave this to future work. Co-evolution did not outperform no-plasticity evolution in any tested condition (two ties at fitness ceilings, one marginally negative result); a stronger test would require harder benchmarks where structural robustness alone is insufficient.

\textbf{Methodological limitations.} The random-RNN sample is small (65 competent networks), and the weight distribution differs from MorphoNAS (uniform vs distance-dependent), introducing a potential confound. We do not compare against alternative optimised encodings (e.g., NEAT~\cite{stanley2002evolving}, HyperNEAT~\cite{stanley2009hypercube}); whether the developmental advantage we identify is specifically morphogenetic or holds for any indirect encoding of comparable expressivity remains open, and is the natural baseline for future work. The plasticity rule (Eq.~\ref{eq:hebb}) is deliberately simple---a single $(\eta, \lambda)$ shared across all synapses---and richer mechanisms (e.g., per-synapse mixed Hebbian and anti-Hebbian rules, neuromodulation~\cite{soltoggio2018born}) may interact differently with developmental topologies.

\textbf{Broader context.} Zador~\cite{zador2019critique} argued that biological brains leverage both genomic encoding and environmental interaction. Shuvaev et~al.~\cite{shuvaev2024encoding} confirmed this computationally, showing that small genomic networks can compress weight matrices by orders of magnitude while retaining substantial task performance. Our findings extend this picture with empirical evidence: the morphogenetic genome provides the compressed architectural scaffold (with an $8.4\times$ competence advantage over random wiring), plasticity provides within-lifetime adaptation, and evolution can discover appropriate plasticity parameters for each task. Across all experiments, the primary determinant of performance remains the developmental topology itself.

\begin{acks}
AI-assisted tools (Claude, Anthropic) were used to write and debug experimental and analysis code under author direction, and to edit and polish the manuscript text. All scientific decisions---including experimental design, data interpretation, and conclusions---are the authors' own. Source code and data are available at \url{https://github.com/ukma-morphonas-lab/MorphoNAS-PL}.
\end{acks}


\end{document}